# Probabilistic Description Logics

Jochen Heinsohn
German Research Center for Artificial Intelligence (DFKI)
Stuhlsatzenhausweg 3
D-66123 Saarbrücken, Germany
email: heinsohn@dfki.uni-sb.de

## Abstract

On the one hand, classical *terminological knowledge representation* excludes the possibility of handling uncertain concept descriptions involving, e.g., "usually true" concept properties, generalized quantifiers, or exceptions. On the other hand, purely numerical *approaches for handling uncertainty* in general are unable to consider terminological knowledge. This paper presents the language $\mathcal{ALCP}$ which is a probabilistic extension of terminological logics and aims at closing the gap between the two areas of research. We present the formal semantics underlying the language $\mathcal{ALCP}$ and introduce the probabilistic formalism that is based on classes of probabilities and is realized by means of probabilistic constraints. Besides inferring implicitly existent probabilistic relationships, the constraints guarantee terminological and probabilistic consistency. Altogether, the new language $\mathcal{ALCP}$ applies to domains where both term descriptions and uncertainty have to be handled.

## 1 INTRODUCTION

Research in knowledge representation led to the development of terminological logics [Nebel, 1990] which originated mainly in Brachman's KL-ONE [Brachman and Schmolze, 1985] and are called *description logics* [Patil et al., 1992] since 1991. In such languages the terminological formalism (*TBox*) is used to represent a hierarchy of terms (*concepts*) that are partially ordered by a subsumption relation: concept $B$ is *subsumed by* concept $A$, if, and only if, the set of $B$'s real world objects is necessarily a subset of $A$'s world objects. In this sense, the semantics of such languages can be based on set theory. Two-place relations (*roles*) are used to describe concepts. In the case of *defined* concepts, restrictions on roles represent both necessary and sufficient conditions. For *primitive* concepts, only necessary conditions are specified. The algorithm called *classifier* inserts new generic concepts at the most specific place in the terminological hierarchy according to the subsumption relation. Work on terminological languages led further to *hybrid* representation systems. Systems like BACK, CLASSIC, LOOM, KANDOR, KL-TWO, KRIS, KRYPTON, MESON, SB-ONE, and YAK (for overview and analyses see [Sigart Bulletin, 1991; Heinsohn et al., 1994]) make use of a separation of terminological and assertional knowledge.

Since, on the one hand, the idea of terminological representation is essentially based on the possibility of *defining* concepts (or at least specifying necessary conditions), the classifier can be employed to draw correct inferences. On the other hand, characterizing domain concepts only *categorically* can lead to problems, especially in domains where certain important properties cannot be used as part of a concept definition. This may happen especially in real world applications where, besides their description, terms can only be characterized as having additional *typical* properties or properties that are, for instance, *usually* true. In the real world such properties often are only *tendencies*. Until now, tendencies as well as differences in these tendencies have not been considered in the framework of terminological logics.

While, as argued above, classical *terminological knowledge representation* excludes the possibility of handling uncertain concept descriptions, purely numerical *approaches for handling uncertainty* (see, e.g., [Kruse et al., 1991]) in general are unable to consider terminological knowledge. The basic idea underlying the formalism presented in this paper is to generalize terminological logics by using probabilistic semantics and in this way to close the gap between the two areas of research.

This paper presents the language $\mathcal{ALCP}$ [Heinsohn, 1993] which is a probabilistic extension of terminological logics and allows one to handle the problems discussed above. First, we briefly introduce $\mathcal{ALC}$ [Schmidt-Schauß and Smolka, 1991], a propositionally complete terminological language containing the logical connectives conjunction, disjunction and negation, as well as role quantification. In Section 3 we extend $\mathcal{ALC}$ by defining the syntax and semantics of *probabilistic conditioning* (p-conditioning), a construct aimed at considering uncertain knowledge sources and based on a statistical interpretation. In Section 4 we introduce the formal model underlying both the terminological and the probabilistic formalism. We further characterize the classes of probabilities induced by a terminology and a set of p-conditionings. As demonstrated in Section 5, a set of consistency requirements have to be met on the basis of terminological and probabilistic knowledge. Moreover, the developed interval-valued probabilistic constraints allow the inference of implicitly existent probabilistic relationships and their quantitative computation. Related work and the conclusions are given in Sections 6 and 7, respectively. The research presented in this paper completes our earlier investigations that introduced first probabilistic constraints [Heinsohn, 1991] and discussed the importance of subsumption computation in the framework of probabilistic knowledge [Heinsohn, 1992].

## 2   THE TERMINOLOGICAL FORMALISM

The basic elements of the terminological language $\mathcal{ALC}$ [Schmidt-Schauß and Smolka, 1991] are concepts and roles (denoting subsets of the domain of interest and binary relations over this domain, respectively). Assume that $\top$ ("top", denoting the entire domain) and $\bot$ ("bottom", denoting the empty set) are concept symbols, that $A$ denotes a concept symbol, and $R$ denotes a role. Then the concepts (denoted by letters $C$ and $D$) of the language $\mathcal{ALC}$ are built according to the abstract syntax rule

$$
\begin{array}{rll}
C, D \rightarrow & A \mid \top \mid \bot & \text{atomic concepts} \\
& C \sqcap D & \text{concept conjunction} \\
& C \sqcup D & \text{concept disjunction} \\
& \neg C & \text{concept negation} \\
& \forall R : C & \text{value restriction} \\
& \exists R : C & \text{existential restriction}
\end{array}
$$

With an introduction to formal semantics of $\mathcal{ALC}$ in mind, we give a translation into set theoretical expressions with $\mathcal{D}$ being the domain of discourse. For that purpose, we define a mapping $\mathcal{E}$ that maps every concept description to a subset of $\mathcal{D}$ and every role to a subset of $\mathcal{D} \times \mathcal{D}$ in the following way:

$$
\begin{aligned}
\mathcal{E}[\top] &= \mathcal{D} \\
\mathcal{E}[\bot] &= \emptyset \\
\mathcal{E}[(C \sqcap D)] &= \mathcal{E}[C] \cap \mathcal{E}[D] \\
\mathcal{E}[(C \sqcup D)] &= \mathcal{E}[C] \cup \mathcal{E}[D] \\
\mathcal{E}[(\neg C)] &= \mathcal{D} \setminus \mathcal{E}[C] \\
\mathcal{E}[(\forall R : C)] &= \{x \in \mathcal{D} \mid \text{for all } y \in \mathcal{D} : \\
&\qquad ((x, y) \in \mathcal{E}[R] \Rightarrow y \in \mathcal{E}[C])\} \\
\mathcal{E}[(\exists R : C)] &= \{x \in \mathcal{D} \mid \text{there exists } y \in \mathcal{D} : \\
&\qquad ((x, y) \in \mathcal{E}[R] \land y \in \mathcal{E}[C])\}
\end{aligned}
$$

Concept descriptions are used to state necessary, or necessary and sufficient conditions by means of specializations "$\sqsubseteq$" or definitions "$\doteq$", respectively. Assuming symbol $A$ and concept description $C$, then "$A \sqsubseteq C$" means the inequality $\mathcal{E}[A] \subseteq \mathcal{E}[C]$, and "$A \doteq C$" means the equation $\mathcal{E}[A] = \mathcal{E}[C]$. A set of well formed concept definitions and specializations forms a *terminology*, if every concept symbol appears at most once on the left hand side and there are no terminological cycles.

**Definition 1** *Let $\mathcal{T}$ be a terminology. The set*

$$\mathrm{mod}(\mathcal{T}) \stackrel{\text{def}}{=} \{\mathcal{E} \mid \mathcal{E} \text{ extension function of } \mathcal{T}\} \quad (1)$$

*is called set of models of $\mathcal{T}$.*

A concept $C_1$ is said to be *subsumed by* a concept $C_2$ in a terminology $\mathcal{T}$, written $C_1 \preceq_\mathcal{T} C_2$, iff the inequality $\mathcal{E}[C_1] \subseteq \mathcal{E}[C_2]$ holds for all extension functions satisfying the equations introduced in $\mathcal{T}$ (i.e., for all $\mathcal{E} \in \mathrm{mod}(\mathcal{T})$).

Terminological languages as $\mathcal{ALC}$ can be usefully applied to *categorical* world knowledge. For instance, we may introduce

**Example 1**

$$
\begin{array}{rcl}
animal & \sqsubseteq & \top \\
flying & \sqsubseteq & \top \\
antarctic\_animal & \sqsubseteq & animal \\
bird & \doteq & animal \sqcap (\forall moves\_by : flying) \\
antarctic\_bird & \doteq & antarctic\_animal \sqcap bird \\
penguin & \sqsubseteq & antarctic\_bird
\end{array}
$$

To characterize the expressiveness of terminological languages, we examine the different relations imaginable between two concept *extensions*, i.e., inclusion, disjointness, and overlapping: Inclusion can be caused by terminological subsumption. For instance, in Example 1 the set of penguins is known to be a subset of the set of antarctic birds. Also disjointness can be a terminological property. For instance, the above language construct "concept negation" used in the expression $C_1 \sqsubseteq C$, $C_2 \doteq (C \sqcap \neg C_1)$ implies $\mathcal{E}[C_1] \cap \mathcal{E}[C_2] = \emptyset$. However, the information of overlapping concept extensions cannot be expressed and used in classical terminological logics. The importance of having such language constructs becomes obvious, if we examine the above birds' taxonomy in more detail: Because of terminological subsumption, the flying property of birds is inherited also to the penguin concept. However, it is well known that concerning this aspect penguins represent a real exception, so that the (categorical) definition of birds seems also to be inadequate: At best *"most* birds move by flying" or are flying objects that are defined to move by flying. It seems to be more suitable to consider generally the "degree of intersection" between the respective concept's extensions and to characterize it using an appropriate technique. The idea behind this generalization is to use probabilistic semantics.

# 3  PROBABILISTIC CONDITIONING

In the following we consider only one representative for equivalent concept expressions (such as $A$, $A \sqcap \top$, $A \sqcap A$). The algebra based on representatives of equivalence classes and on the logical connectives $\sqcap$, $\sqcup$, and $\neg$ is known as *Lindenbaum algebra* of the set $\mathcal{S}$ of concept symbols. We use the symbol $\mathcal{C}$ for the set of concept descriptions. Domain $\mathcal{D}$ is assumed to be finite. As a language construct that takes into account *overlapping* concept extensions, we introduce the notion of *p-conditioning*: the language construct $C_1 \stackrel{[p_l,p_u]}{\to} C_2$ is called p-conditioning, iff $[p_l, p_u]$ is a subrange of real numbers with $0 \leq p_l \leq p_u \leq 1$ and $C_1, C_2 \in \mathcal{C}$. The semantic is defined as follows:

**Definition 2** *An extension function $\mathcal{E}$ over $\mathcal{C}$ satisfies a p-conditioning $C_1 \stackrel{[p_l,p_u]}{\to} C_2$, written $\models_{\mathcal{E}} C_1 \stackrel{[p_l,p_u]}{\to} C_2$, iff*

$$\frac{|\mathcal{E}[C_1 \sqcap C_2]|}{|\mathcal{E}[C_1]|} \in [p_l, p_u] \qquad (2)$$

*holds for concepts $C_1, C_2 \in \mathcal{C}$, $\mathcal{E}[C_1] \neq \emptyset$.*

In case of $p_l = p_u$ we simply write $C_1 \stackrel{p}{\to} C_2$ with $p = p_l = p_u$ instead of $C_1 \stackrel{[p_l,p_u]}{\to} C_2$. From the above it is obvious that we use the *relative cardinality* for interpreting the notion of p-conditioning. For illustrating the meaning of Definition 2, assume that an observer examines the flying ability of birds in more detail. When finishing his study he may have learned that, unlike the model of Example 1, relation *moves_by:flying* holds only for a certain percentage of the birds. The notion of p-conditioning now allows a representation of universal knowledge of statistical kind in a way that maintains the semantics of the roles: the new concept *flying_object* is created with role *moves_by* restricted to the range *flying*. The uncertainty is represented by a p-conditioning stating that "at least 95% of *birds* are *flying_objects* that, by definition, all move by flying". The now more detailed view of the example world leads to the following revision of Example 1:

**Example 2** *(revises Example 1)*

| | | |
|---|---|---|
| animal | $\sqsubseteq$ | $\top$ |
| flying | $\sqsubseteq$ | $\top$ |
| flying_object | $\doteq$ | $\forall moves\_by : flying$ |
| antarctic_animal | $\sqsubseteq$ | animal |
| bird | $\sqsubseteq$ | animal |
| antarctic_bird | $\doteq$ | antarctic_animal $\sqcap$ bird |
| penguin | $\sqsubseteq$ | antarctic_bird |
| bird | $\stackrel{[0.95,1]}{\to}$ | flying_object |
| bird | $\stackrel{0.20}{\to}$ | antarctic_bird |
| penguin | $\stackrel{0}{\to}$ | flying_object |

This demonstrates that set theory is sufficient for a consistent semantic basis on which both terminological and probabilistic language constructs can be interpreted. On this basis, the p-conditioning serves also as a generalization of both "inclusion" and "disjointness" (now appearing as $A \stackrel{1}{\to} B$ and $A \stackrel{0}{\to} B$, respectively).

Example 2 shows not only an adequate representation of the fact that "most (i.e., $\geq 95\%$) birds are flying objects" but also that "20% of the birds are antarctic birds" and "no penguin is a flying object". This directly leads to the question in which way inferences can be drawn on the basis of terminological and probabilistic knowledge to infer implicitly existent relationships. In fact, Example 2 implicitly covers the knowledge, that "at least 75% of antarctic birds are flying objects" and that "at most 5% of birds are penguins", for instance. For this, we first introduce the formal model based on classes of probabilities and then derive the associated probabilistic constraints.

# 4  THE FORMAL MODEL

In concrete application domains, knowledge about uncertain concept relations generally exists only for some pairs of concepts of a terminology—neither directly representable statistical knowledge nor textbook knowledge is complete in this sense. Consequently, the question arises in which way, starting with a set of models restricted with respect to terminology and p-conditionings, one can infer (uncertain) relationships between those pairs of concepts for which p-conditionings are not explicitly introduced. Below we give an answer to this question by defining the sets of *entailed* and *minimal* p-conditionings. The first part of the definition considers the fact that the set of models of a terminology (see equation (1)) is generally refined if p-conditionings are introduced.

**Definition 3** *Let $\mathcal{T}$ be a terminology and $\mathcal{I}$ be a set of p-conditionings. Then*

$$\text{mod}_{\mathcal{T}}(\mathcal{I}) \stackrel{\text{def}}{=} \{\mathcal{E} \mid \models_{\mathcal{E}} \mathcal{I}\} \cap \text{mod}(\mathcal{T}) \qquad (3)$$

*is called the set of models of $\mathcal{I}$ wrt. $\mathcal{T}$.*

$$\text{Th}_{\mathcal{T}}(\mathcal{I}) \stackrel{\text{def}}{=} \{I \mid \text{for all } \mathcal{E} \in \text{mod}_{\mathcal{T}}(\mathcal{I}) : \models_{\mathcal{E}} I\} \qquad (4)$$

*is called the set of entailed p-conditionings wrt. $\mathcal{I}$ and $\mathcal{T}$.*

$$\min{}_{\mathcal{T}}(\mathcal{I}) \stackrel{\text{def}}{=} \bigcup_{C,D \in \mathcal{C}} \{C \stackrel{R_{\min}}{\to} D \mid R_{\min} = \bigcap_{R : C \stackrel{R}{\to} D \in \text{Th}_{\mathcal{T}}(\mathcal{I})} R\} \qquad (5)$$

*is called the set of minimal p-conditionings wrt. $\mathcal{I}$ and $\mathcal{T}$.*

These definitions—especially the set defined in (5)—describe a formal model that characterizes the computation of p-conditionings introduced not explicitly and the further refinement of p-conditionings that are known. Note that both sets (4) and (5) contain p-conditionings for *all* pairs of concepts. Further note that (4) generally is of infinite size: For instance, if a p-conditioning $C_1 \stackrel{[p_1,p_2]}{\to} C_2$ is satisfied by all extension functions in $\mathrm{mod}_{\mathcal{T}}(\mathcal{I})$, then all p-conditionings with ranges that cover $[p_1, p_2]$ are also satisfied, i.e.,

$$\left(C_1 \stackrel{[p_1,p_2]}{\to} C_2\right) \in \mathrm{Th}_{\mathcal{T}}(\mathcal{I}) \Rightarrow \left(C_1 \stackrel{[q_1,q_2]}{\to} C_2\right) \in \mathrm{Th}_{\mathcal{T}}(\mathcal{I})$$

holds for all $q_1, q_2 \in [0, 1]$, $q_1 \leq p_1$, $p_2 \leq q_2$. While $\mathrm{Th}_{\mathcal{T}}(\mathcal{I})$ generally is of infinite size, set $\mathrm{min}_{\mathcal{T}}(\mathcal{I})$ defined in (5) contains exactly one p-conditioning for one pair of concepts.

**Definition 4** *a) A set $\mathcal{I}$ of p-conditionings is called consistent wrt. $\mathcal{T}$, iff $\mathrm{mod}_{\mathcal{T}}(\mathcal{I}) \neq \emptyset$ holds.*
*b) A concept $C_1$ is said to be subsumed by a concept $C_2$ wrt. $\mathcal{T}$ and $\mathcal{I}$, written $C_1 \preceq_{\mathcal{T},\mathcal{I}} C_2$, iff the inequality $\mathcal{E}[C_1] \subseteq \mathcal{E}[C_2]$ holds for all extension functions $\mathcal{E} \in \mathrm{mod}_{\mathcal{T}}(\mathcal{I})$.*

It can be shown that all minimal sets $R_{\min}$ of real numbers defined in (5) form ranges as it is the case for explicitly introduced p-conditionings. This is due to the convexity property of those probability classes that are induced by terminological axioms and p-conditionings over the set of atomic concept expressions. In the following we focus on this aspect.

In addition to the symbol $\mathcal{C}$ for the set of concept descriptions we use $\mathcal{C}^A$ for the set of *atomic concept expressions* (i.e., the atoms of the Lindenbaum algebra). Atomic concept expressions are of the form $B_1 \sqcap B_2 \sqcap \ldots \sqcap B_m$, where $B_i$ is either a concept symbol $A$ or the negation $\neg A$ of a symbol. The relation $\mathcal{C}^A \subseteq \mathcal{C}$ holds. A first simple observation is that for every extension function $\mathcal{E} \in \mathrm{mod}_{\mathcal{T}}(\mathcal{I})$ the set of extensions of the elements in $\mathcal{C}^A$ forms a partition of $\mathcal{D}$. A direct consequence of this observation is that every extension function $\mathcal{E}$ uniquely determines a probability over $\mathcal{C}^A$.

**Proposition 1** *Let $\mathcal{T}$ be a terminology and $\mathcal{I}$ be a consistent set of p-conditionings. Further, let $\mathcal{E} \in \mathrm{mod}_{\mathcal{T}}(\mathcal{I})$ be an extension function for which*

$$\models_{\mathcal{E}} \top \stackrel{p_i}{\to} C_i^-, \quad p_i \stackrel{\mathrm{def}}{=} \frac{|\mathcal{E}[C_i^-]|}{|\mathcal{D}|} \text{ for all } C_i^- \in \mathcal{C}^A \quad (6)$$

*holds. Then the real-valued set function $P_{\mathcal{E}}$ defined by*

$$P_{\mathcal{E}} : 2^{\mathcal{C}^A} \to [0,1], \; P_{\mathcal{E}}(\{C_i^-\}) \stackrel{\mathrm{def}}{=} p_i, \; C_i^- \in \mathcal{C}^A \quad (7)$$

*is a probability function over $\mathcal{C}^A$.*

**Proof:** Let $\mathcal{E}$ be an extension function for which (6) holds. Since $\mathcal{E}$ induces a partition of $\mathcal{D}$ and because of the semantic (2) of p-conditionings we derive

$$P_{\mathcal{E}}(\mathcal{C}^A) = 1,$$
$$P_{\mathcal{E}}(D_i) \geq 0 \text{ for all } D_i \subseteq \mathcal{C}^A,$$
$$P_{\mathcal{E}}(D_i \cup D_j) = P_{\mathcal{E}}(D_i) + P_{\mathcal{E}}(D_j) \text{ if } D_i \cap D_j = \emptyset.$$

Consequently, $P_{\mathcal{E}}$ is a probability function over $\mathcal{C}^A$. ∎

**Example 3** *Assume the set*

$$\mathcal{T} = \{A \sqsubseteq \top, B \sqsubseteq \top, C \doteq A \sqcap B\}$$

*of terminological axioms. From $\mathcal{S} = \{\top, A, B\}$ we obtain $\mathcal{C}^A = \{\neg A \sqcap \neg B, \neg A \sqcap B, A \sqcap B, A \sqcap \neg B\}$. Then, $\mathcal{E}$ and $\mathcal{D}$ with $|\mathcal{D}| = 100, |\mathcal{E}[A]| = 40, |\mathcal{E}[B]| = 20, |\mathcal{E}[A \sqcap B]| = 10$ induce a probability function*

$$P_{\mathcal{E}} : \neg A \sqcap \neg B \mapsto 0.5$$
$$\neg A \sqcap B \mapsto 0.1$$
$$A \sqcap B \mapsto 0.1$$
$$A \sqcap \neg B \mapsto 0.3.$$

Note that every concept can be represented as a disjunction of atomic concept expressions, i.e., for every concept expression $C \in \mathcal{C}$ there exists a subset $D \subseteq \mathcal{C}^A$ of atoms such that $C = \bigsqcup D$. In this way $P_{\mathcal{E}}$ can be extended to concept expressions. In particular,

$$P_{\mathcal{E}}(\top) = P_{\mathcal{E}}(\bigsqcup \mathcal{C}^A) = 1,$$
$$P_{\mathcal{E}}(C) \geq 0 \text{ for all } C \in \mathcal{C},$$
$$P_{\mathcal{E}}(C_i \sqcup C_j) = P_{\mathcal{E}}(C_i) + P_{\mathcal{E}}(C_j) \text{ if } C_i \sqcap C_j \preceq_{\mathcal{T},\mathcal{I}} \bot$$

hold, so that $P_{\mathcal{E}}$ defined by

$$P_{\mathcal{E}} : \mathcal{C} \to [0,1],$$
$$P_{\mathcal{E}}(C) \stackrel{\mathrm{def}}{=} \sum_{C^- : C^- \in D, D \subseteq \mathcal{C}^A, C = \bigsqcup D} P_{\mathcal{E}}(\{C^-\})$$

is a probability function over $\mathcal{C}$.

Example 3 shows that, assuming complete knowledge of domain $\mathcal{D}$ and of the involved cardinalities, a probability function $P_{\mathcal{E}}$ over $\mathcal{C}^A$ is induced by the extension function $\mathcal{E}$. However, it is generally more realistic to assume less complete knowledge and cardinalities that are rather relative. Consequently, the set $\mathrm{mod}_{\mathcal{T}}(\mathcal{I})$ generally contains more than one element, so that a *class of probabilities* is induced by a terminology and a set of p-conditionings. The most general set of all probabilities over $\mathcal{C}^A$ is defined by

$$\mathcal{M} \stackrel{\mathrm{def}}{=} \{(p_1,\ldots,p_n) \in \Re^n \mid p_1 + \ldots + p_n = 1,$$
$$p_i \geq 0 \text{ for all } 1 \leq i \leq n\},$$

with $n = 2^m$. Without any knowledge about terminology and p-conditionings the set $\mathcal{M}$ characterizes the status of *complete ignorance*. On the other hand, for a particular extension function $\mathcal{E}$, the set $\mathcal{M}$ consists of exactly one point in the $n$-dimensional space $[0,1]^n$. In the case of the given terminology $\mathcal{T}$ and p-conditionings $\mathcal{I}$, by (3) a set $\mathrm{mod}_{\mathcal{T}}(\mathcal{I})$ of extension functions and also a set

$$\mathcal{M}_{\mathcal{T},\mathcal{I}} \stackrel{\mathrm{def}}{=} \{(p_1,\ldots,p_n) \in \mathcal{M} \mid \text{exists } \mathcal{E} \in \mathrm{mod}_{\mathcal{T}}(\mathcal{I}) :$$
$$p_j = P_{\mathcal{E}}(\{C_j^-\}), j = 1,\ldots,n\}$$

of probabilities are defined. $\mathcal{M}_{\mathcal{T},\mathcal{I}}$ corresponds to the set of probabilities in $\mathcal{M}$ that are compatible with $\mathcal{T}$ and all p-conditionings in $\mathcal{I}$.

**Proposition 2** *For every consistent $\mathcal{T}$ and $\mathcal{I}$ the set $\mathcal{M}_{\mathcal{T},\mathcal{I}} = \bigcap_{I \in \mathcal{I}} \mathcal{M}_{\mathcal{T},\{I\}}$ is convex.*

**Proof:** We show that p-conditionings can be represented as linear combinations of atomic concept expressions. Assume p-conditioning $I : C_1 \stackrel{[q_l,q_u]}{\to} C_2$, $C_1, C_2 \in \mathcal{C}$. $I$ can be rewritten as

$$\begin{aligned} P_{\mathcal{E}}(C_1 \sqcap C_2) &\geq q_l \cdot P_{\mathcal{E}}(C_1), \\ q_u \cdot P_{\mathcal{E}}(C_1) &\geq P_{\mathcal{E}}(C_1 \sqcap C_2), \end{aligned} \quad (8)$$

if $P_{\mathcal{E}}(C_1) \neq 0$. With $\mathcal{C}^A$ as the set of atoms of the Lindenbaum algebra, the concept expressions $C_1 \sqcap C_2$ and $C_1$ can be substituted by disjunctions of atomic expressions:

$$C_1 \sqcap C_2 = \bigsqcup_{j=1}^{k} C_{1,j}^{-}, \quad C_1 = \bigsqcup_{j=1}^{l} C_{1,j}^{-}, \quad k \leq l.$$

Therefore, both inequalities (8) can be represented by expressions of the general form

$$\sum_{i=1}^{n} x_i p_i \geq \sum_{i=1}^{n} y_i p_i, \text{ with } p_i = P_{\mathcal{E}}(\{C_i^{-}\}),\ C_i^{-} \in \mathcal{C}^A$$

where $x_i$ ($y_i$) is 0 if the corresponding concept disjunction does not contain $C_i^-$, and $q_u$ ($q_l$) or 1 otherwise. Making use of $z_i = x_i - y_i$ we derive the general representation $\sum_{i=1}^n z_i p_i \geq 0$ of the inequalities and the set

$$\begin{aligned}\mathcal{M}_{\mathcal{T},\{I\}} &= \{(p_1,\ldots,p_n) \in \mathcal{M} \mid C_1 \stackrel{[q_l,q_u]}{\to} C_2\} \\ &= \{(p_1,\ldots,p_n) \in \mathcal{M} \mid \sum_{i=1}^{n} z_i p_i \geq 0\}\end{aligned}$$

with appropriate values $z_i$. Using a geometric interpretation, each inequality defines a (convex) hyperplane. Since the intersection of convex regions is known to be always convex, $\mathcal{M}_{\mathcal{T},\mathcal{I}} = \bigcap_{I \in \mathcal{I}} \mathcal{M}_{\mathcal{T},\{I\}}$ is convex, too. ∎

This convexity property represents a sufficient condition for the existence of the intersections used in (5) and of the probabilistic constraints derived below.[1]

## 5 PROBABILISTIC CONSTRAINTS

In the following, we focus on *probabilistic constraints* corresponding to the formal model introduced above. They are *locally* defined and therefore *context-related*, and they *derive* and *refine* p-conditionings and check in this way the consistency of the knowledge base. The following simple constraints characterize the relations between subsumption and p-conditioning ((9) and (10)), state that non-trivial reflexive p-conditionings do not exist ((11)), and focus on the role of disjointness ((12) and (13)).

**Proposition 3** *Assume consistent sets $\mathcal{T}$ and $\mathcal{I}$, and abbreviation $\mathcal{J} = \min_{\mathcal{T}}(\mathcal{I})$. For all concepts $C, D \in \mathcal{C} \setminus \{\bot\}$:*

$$D \preceq_{\mathcal{T},\mathcal{I}} C \Leftrightarrow \left(D \stackrel{1}{\to} C\right) \in \mathcal{J} \quad (9)$$

$$(C \sqcap D) \preceq_{\mathcal{T},\mathcal{I}} \bot \Leftrightarrow \left(D \stackrel{0}{\to} C\right) \in \mathcal{J} \quad (10)$$

$$\left(C \stackrel{[p_l,p_u]}{\to} C\right) \in \mathcal{J} \Rightarrow p_l = p_u = 1 \quad (11)$$

$$\left(C \stackrel{0}{\to} D\right) \in \mathcal{J} \Leftrightarrow \left(D \stackrel{0}{\to} C\right) \in \mathcal{J} \quad (12)$$

$$\{C \stackrel{[p_l,p_u]}{\to} D, D \stackrel{[q_l,q_u]}{\to} C\} \subseteq \mathcal{J}$$
$$\Rightarrow (p_l > 0 \Leftrightarrow q_l > 0) \quad (13)$$

**Proof:** The proof of (9) is based on Definition 4 and on equations (4) and (5):

$$\begin{aligned}D \preceq_{\mathcal{T},\mathcal{I}} C &\Leftrightarrow \mathcal{E}[D] \subseteq \mathcal{E}[C] \text{ for all } \mathcal{E} \in \mathrm{mod}_{\mathcal{T}}(\mathcal{I}) \\ &\Leftrightarrow \frac{|\mathcal{E}[C \sqcap D]|}{|\mathcal{E}[D]|} = 1 \text{ for all } \mathcal{E} \in \mathrm{mod}_{\mathcal{T}}(\mathcal{I}) \\ &\Leftrightarrow \left(D \stackrel{1}{\to} C\right) \in \mathrm{Th}_{\mathcal{T}}(\mathcal{I}) \\ &\Leftrightarrow \left(D \stackrel{1}{\to} C\right) \in \min_{\mathcal{T}}(\mathcal{I})\end{aligned}$$

The other proofs are obtained by analogy. ∎

In the rest of this paper we restrict ourselves to *triangular cases* that take into account three concept expressions and allow the inference of minimal p-conditionings. Note that both of the following propositions examine the most general triangular case that exists for sets of *primitive* concepts.[2] If a subsumption relation between concepts is known, the corresponding p-conditioning has to have the range $[1,1]$ (compare (9)).

**Proposition 4** *Assume concepts $A$, $B$, $C$, and p-conditionings*

$$\mathcal{I} = \{A \stackrel{[p_l,p_u]}{\to} C, A \stackrel{[q_l,q_u]}{\to} B, B \stackrel{[q'_l,q'_u]}{\to} A, C \stackrel{[p'_l,p'_u]}{\to} A\}.$$

*The minimal p-conditioning $B \stackrel{R_{\min}}{\to} C \in \min_{\mathcal{T}}(\mathcal{I})$ derivable on the basis of this knowledge has the range $R_{\min} = [r_l, r_u]$ with*

$$r_l = \begin{cases} \dfrac{q'_l}{q_l} \cdot \max(0, q_l + p_l - 1) & \text{if } q_l \neq 0, \\ q'_l & \text{if } q_l = 0, p_l = 1, \\ 0 & \text{otherwise,} \end{cases}$$

---

[1] Note that *comparative assertions* such as "the percentage of birds that fly is greater than the percentage of dogs that bark" may lead to non-convexity. Therefore, we excluded such qualitative language constructs from $\mathcal{ALCP}$.

[2] The proofs are omitted here for lack of space (see [Heinsohn, 1993]). A condensed English version of the thesis containing proofs as well is in preparation.

$$r_u = \begin{cases} \min\left(1,\ 1 - q'_l + p_u \cdot \dfrac{q'_l}{q_l},\ \dfrac{p_u}{p'_l} \cdot \dfrac{q'_u}{q_l},\right. \\ \qquad \dfrac{p_u}{p'_l} \cdot \dfrac{q'_u}{q_l} \cdot (1 - p'_l) + q'_u, \\ \qquad \left.\dfrac{p_u}{p'_l \cdot (q_l - p_u) + p_u}\right) & \text{if } p'_l \neq 0, q_l \neq 0, \\ \min\left(1,\ 1 - q'_l + p_u \cdot \dfrac{q'_l}{q_l}\right) & \text{if } p'_l = 0, q_l \neq 0, \\ q'_u & \text{if } p'_l = 1, q_l \neq 0, \\ 1 - q'_l & \text{if } p_u = 0, q_l = 0, \\ 1 & \text{otherwise.} \end{cases}$$

Note that for the considered set of at most four known p-conditionings, Proposition 4 leads to the minimal p-conditioning $B \overset{R_{\min}}{\to} C$. The associated constraints already take into account the possible consistent range of the unknown p-conditioning $C \overset{R}{\to} B$. However, if five ranges are known also the following constraint that can be simply derived from Bayes rule has to be applied to guarantee local completeness. The reason for this additional constraint is that the consistent range which can be derived for the p-conditioning $C \overset{R}{\to} B$ may lead to a refinement of the range that is explicitly given (and vice versa).

**Proposition 5** *Assume concepts A, B, C, and p-conditionings*

$$\mathcal{I} = \{A \overset{[p_l,p_u]}{\to} C, A \overset{[q_l,q_u]}{\to} B, B \overset{[q'_l,q'_u]}{\to} A, C \overset{[p'_l,p'_u]}{\to} A,\\ C \overset{[r'_l,r'_u]}{\to} B,\ p'_l \neq 0, q_l \neq 0\}.$$

*The minimal p-conditioning $B \overset{R_{\min}}{\to} C \in \min_{\mathcal{T}}(\mathcal{I})$ derivable on the basis of this knowledge has the range*

$$R_{\min} \subseteq \left[r'_l \cdot \dfrac{p_l}{p'_u} \cdot \dfrac{q'_l}{q_u},\ r'_u \cdot \dfrac{p_u}{p'_l} \cdot \dfrac{q'_u}{q_l}\right]. \quad (14)$$

The following examples visualize the "behaviour" of the probabilistic constraints for some special cases. In particular, Example 5 shows that the constraints also apply to the situation that has been discussed at the end of Section 3:

**Example 4** *In the first situation below, we consider given point values. In this case, only constraint (14) leads to a refinement. The incoming and computed ranges are shown in rows (i) and (ii), respectively:*

|      | $r'$:        | $r$:         | $q'$:        |
|------|--------------|--------------|--------------|
| (i)  | [0.00 1.00]  | [0.50 0.50]  | [0.50 0.50]  |
| (ii) | [0.50 0.50]  | [0.50 0.50]  | [0.50 0.50]  |

|  | $q$:        | $p$:        | $p'$:       |
|--|-------------|-------------|-------------|
|  | [0.50 0.50] | [0.50 0.50] | [0.50 0.50] |
|  | [0.50 0.50] | [0.50 0.50] | [0.50 0.50] |

*As shown in the table, for the completely unspecified variable $r'$ the minimal range $[0.5, 0.5]$ can be derived.*

*In the following case the constraints specified in both Propositions, 4 and 5, apply. In particular, table*

|       | $r'$:        | $r$:         | $q'$:        |
|-------|--------------|--------------|--------------|
| (i)   | [1.00 1.00]  | [0.50 0.50]  | [0.50 1.00]  |
| (ii)  | [1.00 1.00]  | [0.50 0.50]  | [0.50 0.75]  |
| (iii) | [1.00 1.00]  | [0.50 0.50]  | [0.50 0.75]  |

|  | $q$:        | $p$:        | $p'$:       |
|--|-------------|-------------|-------------|
|  | [0.50 0.50] | [0.10 1.00] | [0.50 0.50] |
|  | [0.50 0.50] | [0.10 0.50] | [0.50 0.50] |
|  | [0.50 0.50] | [0.17 0.25] | [0.50 0.50] |

*shows in row (ii) that first the ranges of $q'$ and $p$ are refined by applying the constraint of Proposition 4. After that $p$ is again refined by constraint (14) as shown in row (iii).*

**Example 5** *On the basis of*

$$\text{antarctic\_bird} \preceq_{\mathcal{T},\mathcal{I}} \text{bird},$$
$$\text{bird} \overset{0.20}{\to} \text{antarctic\_bird},$$
$$\text{bird} \overset{[0.95,1]}{\to} \text{flying\_object}$$

*we derive* $\text{antarctic\_bird} \overset{[0.75,1]}{\to} \text{flying\_object}$. *Similarly,*

$$\text{penguin} \preceq_{\mathcal{T},\mathcal{I}} \text{bird},$$
$$\text{penguin} \overset{0}{\to} \text{flying\_object},$$
$$\text{bird} \overset{[0.95,1]}{\to} \text{flying\_object}$$

*allows to infer p-conditioning* $\text{bird} \overset{[0,0.05]}{\to} \text{penguin}$.

Propositions 4 and 5 cover several interesting special cases such as *chaining*, i.e.,

$$C_3 \preceq_{\mathcal{T},\mathcal{I}} C_2 \preceq_{\mathcal{T},\mathcal{I}} C_1, C_1 \overset{[p_l,p_u]}{\to} C_2, p_l \neq 0, C_1 \overset{[q_l,q_u]}{\to} C_3$$

implies

$$C_2 \overset{R_{\min}}{\to} C_3 \quad \text{with} \quad R_{\min} = \left[\dfrac{q_l}{p_u}, \min\left(1, \dfrac{q_u}{p_l}\right)\right],$$

and *monotonicity*

$$\{C_2 \preceq_{\mathcal{T},\mathcal{I}} C_3, C_1 \overset{[q_l,q_u]}{\to} C_2\} \Rightarrow C_1 \overset{[q_l,1]}{\to} C_3,$$
$$\{C_2 \preceq_{\mathcal{T},\mathcal{I}} C_3, C_1 \overset{[p_l,p_u]}{\to} C_3\} \Rightarrow C_1 \overset{[0,p_u]}{\to} C_2.$$

While the above two propositions examine situations, in which only *primitive* concepts are involved, we show below that in the case of logically interrelated concepts probabilistic constraints have to be further strengthened to guarantee the minimality of ranges. In particular, concept negation, conjunction, and disjunction are considered.

**Proposition 6** *Assume concepts $A, B, C \in \mathcal{C} \setminus \{\bot\}$ and that $\mathcal{J}$ denotes the set $\min_{\mathcal{T}}(\mathcal{I})$. Then*

$$\left(B \overset{[p_l,p_u]}{\to} \neg B\right) \in \mathcal{J} \Rightarrow p_u = 0$$

$$\left(A \overset{[p_l,p_u]}{\to} B\right) \in \mathcal{J} \Leftrightarrow \left(A \overset{[1-p_u,1-p_l]}{\to} \neg B\right) \in \mathcal{J}$$

$$\left(A \overset{[p_l,p_u]}{\to} C\right) \in \mathcal{J} \quad \Leftrightarrow \quad \left(A \overset{[p_l,p_u]}{\to} A \sqcap C\right) \in \mathcal{J}$$

$$\left(A \overset{[p_l,p_u]}{\to} A \sqcap B\right) \in \mathcal{J} \quad \Leftrightarrow \quad \left(A \overset{[1-p_u,1-p_l]}{\to} A \sqcap \neg B\right) \in \mathcal{J}$$

The main advantage of examining local *triangular cases* is that "most" of the inconsistencies are discovered early and can be taken into account in just the *current context* of the three concepts involved. Further, not as yet known p-conditionings can be generated and the associated probability ranges can be stepwise refined. In the general case, testing probabilistic consistency leads, for every p-conditioning, to a successive computing of the intersections of probability ranges derived from different local examinations.

## 6 RELATED WORK

The importance of providing an integration of both term classification and uncertainty representation[3] was recently emphasized in some publications. However, they differ from each other and also from our work. For example, Yen [1991] proposes an extension of term subsumption languages to fuzzy logic that aims at representing and handling vague concepts. His approach generalizes a subsumption test algorithm for dealing with the notion of vagueness and imprecision. Since the language $\mathcal{ALCP}$ aims at modeling uncertainty, it already differs from Yen's proposal in its general objectives. Saffiotti [1990] presents a hybrid framework for representing epistemic uncertainty. His extension allows one to model uncertainty about categorical knowledge, e.g., to express one's belief on quantified statements such as "I am fairly (80%) sure that all birds fly". Note the difference from "I am sure that 80% of birds fly", which is modeled in this paper and requires a different formal basis. The work of Bacchus [1990] is important because he not only explores the question of how far one can go using *statistical* knowledge but also presents LP, a logical formalism for representing and reasoning with statistical knowledge. In spite of being closely related to our work, Bacchus does not provide a deep discussion of conditionals and the associated local consistency requirements.

From the viewpoint of interval-valued probabilistic constraints, our work has been influenced by the early paper [Dubois and Prade, 1988], where first probabilistic constraints were presented. In this framework the system INFERNO [Quinlan, 1983] also has to be mentioned since it is based on the intuition of computing "maximally consistent ranges" underlying the language $\mathcal{ALCP}$. Probabilistic constraints that are related to our work were independently developed by Thöne et al. [1992] in the context of deductive databases and by Amarger et al. [Amarger et al., 1991;

[3]Brachman [1990] considers "probability and statistics" as one of the "potential highlights" in knowledge representation.

Dubois et al., 1992]. Thöne et al. presented an improved upper bound for the interval-valued situation discussed in Proposition 4. Their result allowed us to refine our earlier constraints and has been adopted in this paper. One basic difference to the work on constraints discussed above is that the terminological formalism of $\mathcal{ALCP}$ allows for subsumption computation and for correctly handling logically interrelated concepts. One consequence is that the integrated terminological and probabilistic formalism is able to apply *refined* constraints if necessary [Heinsohn, 1991; Heinsohn, 1992].

While this paper focuses mainly on terminological and probabilistic aspects of *generic* knowledge, the consideration of *assertions* would mean the ability to draw inferences about "probabilistic memberships" of instances and associated belief values. A corresponding extension of $\mathcal{ALCP}$ that is based on probability distributions over both, domains and worlds, is described in [Heinsohn, 1993]. If we enlarge our discussion of related work to this borderline between statistical and belief knowledge and to the question how statistical knowledge can be used to derive *beliefs*, other work has to be mentioned, too: While Bacchus et al. [1992] and Shastri [1989] *examine this question in the general* frameworks of first-order logic and semantic networks, respectively, in [Jäger, 1994] an extension of terminological logics is presented. While Jäger employs cross entropy minimization to derive beliefs, the assertional formalism of $\mathcal{ALCP}$ makes use of the maximally consistent ranges derived in the generic knowledge base. Finally, it is worth pointing out that the constraint interpretation used in this paper is only one of several conceivable ways of integrating probabilities with terminological logics.

## 7 CONCLUSIONS

We have proposed the language $\mathcal{ALCP}$ which is a probabilistic extension of terminological logics. The knowledge, that $\mathcal{ALCP}$ allows us to handle, includes *terminological knowledge* covering term descriptions and *uncertain knowledge* about (not generally true) concept properties. For this purpose, the notion of *probabilistic conditioning* based on a statistical interpretation has been introduced. The developed formal framework for terminological and probabilistic language constructs has been based on *classes of probabilities* that offer a modeling of *ignorance* as one special feature. *Probabilistic constraints* allow the context-related generation and refinement of p-conditionings and check the consistency of the knowledge base. It has been shown that the results of the constraints essentially depend on the correctness of the terminology which is guaranteed by the subsumption algorithm. More details about the language $\mathcal{ALCP}$, the formal framework, the associated interval-valued constraints, proofs, and other related work can be found in [Heinsohn, 1993]. There, an extension for *assertional knowledge* is also offered.

318  Heinsohn

Acknowledgements

I would like to thank Bernhard Nebel and the anonymous referees for their valuable comments on earlier versions of this paper. This research was supported by the German Ministry for Research and Technology (BMFT) under contracts ITW 8901 8 and ITW 9400 as part of the projects WIP and PPP.

References

[Amarger et al., 1991] S. Amarger, D. Dubois, and H. Prade. Imprecise quantifiers and conditional probabilities. In Kruse and Siegel [1991], pages 33–37.

[Bacchus et al., 1992] F. Bacchus, A. Grove, J. Y. Halpern, and D. Koller. From statistics to beliefs. In *Proceedings of the 10th National Conference of the American Association for Artificial Intelligence*, pages 602–608, San Jose, Cal., 1992.

[Bacchus, 1990] F. Bacchus. *Representing and Reasoning With Probabilistic Knowledge*. MIT Press, Cambridge, Mass., 1990.

[Brachman and Schmolze, 1985] R. J. Brachman and J. G. Schmolze. An overview of the KL-ONE knowledge representation system. *Cognitive Science*, 9(2):171–216, 1985.

[Brachman, 1990] R. J. Brachman. The future of knowledge representation. In *Proceedings of the 8th National Conference of the American Association for Artificial Intelligence*, pages 1082–1092, Boston, Mass., 1990.

[Dubois and Prade, 1988] D. Dubois and H. Prade. On fuzzy syllogisms. *Computational Intelligence*, 4(2):171–179, May 1988.

[Dubois et al., 1992] D. Dubois, H. Prade, L. Godo, and R. L. de Mantaras. A symbolic approach to reasoning with linguistic quantifiers. In *Proceedings of the 8th Conference on Uncertainty in Artificial Intelligence*, pages 74–82, Stanford, Cal., July 1992.

[Heinsohn et al., 1994] J. Heinsohn, D. Kudenko, B. Nebel, and H.-J. Profitlich. An empirical analysis of terminological representation systems. *Artificial Intelligence Journal*, 1994. Accepted for publication. A preliminary version is available as DFKI Research Report RR-92-16.

[Heinsohn, 1991] J. Heinsohn. A hybrid approach for modeling uncertainty in terminological logics. In Kruse and Siegel [1991], pages 198–205.

[Heinsohn, 1992] J. Heinsohn. ALCP – an integrated framework to terminological and noncategorical knowledge. In *Proceedings of the 4th International Conference IPMU'92*, pages 493–496, Palma de Mallorca, Spain, July 6–10, 1992.

[Heinsohn, 1993] J. Heinsohn. *ALCP – Ein hybrider Ansatz zur Modellierung von Unsicherheit in terminologischen Logiken*. PhD thesis, Universität des Saarlandes, Saarbrücken, June 1993.

[Jäger, 1994] M. Jäger. Probabilistic reasoning in terminological logics. In J. Doyle, E. Sandewall, and P. Torasso, editors, *Principles of Knowledge Representation and Reasoning: Proceedings of the 4th International Conference*, Bonn, Germany, May 1994. Morgan Kaufmann. To appear.

[Kruse and Siegel, 1991] R. Kruse and P. Siegel, editors. *Symbolic and Quantitative Approaches to Uncertainty, Proceedings of the European Conference ECSQAU*. Lecture Notes in Computer Science 548. Springer, Berlin, Germany, 1991.

[Kruse et al., 1991] R. Kruse, E. Schwecke, and J. Heinsohn. *Uncertainty and Vagueness in Knowledge Based Systems: Numerical Methods*. Series Artificial Intelligence. Springer, Germany, 1991.

[Nebel, 1990] B. Nebel. *Reasoning and Revision in Hybrid Representation Systems*. Lecture Notes in Computer Science, Volume 422. Springer, Berlin, Germany, 1990.

[Patil et al., 1992] R. S. Patil, R. E. Fikes, P. F. Patel-Schneider, D. McKay, T. Finin, T. Gruber, and R. Neches. The DARPA knowledge sharing effort: Progress report. In B. Nebel, W. Swartout, and C. Rich, editors, *Principles of Knowledge Representation and Reasoning: Proceedings of the 3rd International Conference*, pages 777–788, Cambridge, MA, October 1992. Morgan Kaufmann.

[Quinlan, 1983] J. R. Quinlan. INFERNO, a cautious approach to uncertain inference. *Computer Journal*, 26(3):255–269, 1983.

[Saffiotti, 1990] A. Saffiotti. A hybrid framework for representing uncertain knowledge. In *Proceedings of the 8th National Conference of the American Association for Artificial Intelligence*, pages 653–658, Boston, Mass., 1990.

[Schmidt-Schauß and Smolka, 1991] M. Schmidt-Schauß and G. Smolka. Attributive concept descriptions with complements. *Artificial Intelligence Journal*, 48(1):1–26, 1991.

[Shastri, 1989] L. Shastri. Default reasoning in semantic networks: A formalization of recognition and inheritance. *Artificial Intelligence Journal*, 39:283–355, 1989.

[Sigart Bulletin, 1991] *SIGART Bulletin: Special Issue on Implemented Knowledge Representation and Reasoning Systems*, volume 2(3). ACM Press, 1991.

[Thöne et al., 1992] H. Thöne, U. Güntzer, and W. Kießling. Towards precision of probabilistic bounds propagation. In *Proceedings of the 8th Conference on Uncertainty in Artificial Intelligence*, pages 315–322, Stanford, Cal., July 1992.

[Yen, 1991] J. Yen. Generalizing term subsumption languages to fuzzy logic. In *Proceedings of the 12th International Joint Conference on Artificial Intelligence*, Sydney, Australia, 1991.